\documentclass[acmlarge,screen,authorversion]{acmart}

\AtBeginDocument{%
  \providecommand\BibTeX{{%
    \normalfont B\kern-0.5em{\scshape i\kern-0.25em b}\kern-0.8em\TeX}}}

\setcopyright{acmlicensed}
\acmYear{2021} \acmVolume{1} \acmNumber{1} \acmArticle{1} \acmMonth{1} \acmPrice{15.00}\acmDOI{10.1145/3458754}

\acmPrice{15.00}
\acmISBN{978-1-4503-XXXX-X/18/06}

\usepackage{adjustbox}

\usepackage{subcaption}
\begin{document}

\title{Domain-Specific Language Model Pretraining for Biomedical Natural Language Processing}


\author{Yu Gu}
\authornotemark[1]
\email{Aiden.Gu@microsoft.com}

\author{Robert Tinn}
\authornote{These authors contributed equally to this research.}
\email{Robert.Tinn@microsoft.com}

\author{Hao Cheng}
\authornotemark[1]
\email{chehao@microsoft.com}

\author{Michael Lucas}
\email{Michael.Lucas@microsoft.com}

\author{Naoto Usuyama}
\email{naotous@microsoft.com}

\author{Xiaodong Liu}
\email{xiaodl@microsoft.com}

\author{Tristan Naumann}
\email{tristan@microsoft.com}
\orcid{0000-0003-2150-1747}

\author{Jianfeng Gao}
\email{jfgao@microsoft.com}

\author{Hoifung Poon}
\email{hoifung@microsoft.com}

\affiliation{%
  \institution{Microsoft Research}
  \streetaddress{One Microsoft Way}
  \city{Redmond}
  \state{WA}
  \postcode{98052}}

\renewcommand{\shortauthors}{Gu, Tinn, Cheng, et al.}

\newcommand{\eat}[1]{\ignorespaces}

\begin{abstract}

Pretraining large neural language models, such as BERT, has led to impressive gains on many natural language processing (NLP) tasks.
However, most pretraining efforts focus on general domain corpora, such as newswire and Web.
A prevailing assumption is that even domain-specific pretraining can benefit by starting from general-domain language models. 
In this paper, we challenge this assumption by showing that for domains with abundant unlabeled text, such as biomedicine, pretraining language models from scratch results in substantial gains over continual pretraining of general-domain language models.
To facilitate this investigation, we compile a comprehensive biomedical NLP benchmark from publicly-available datasets.
Our experiments show that domain-specific pretraining serves as a solid foundation for a wide range of biomedical NLP tasks, leading to new state-of-the-art results across the board.
Further, in conducting a thorough evaluation of modeling choices, both for pretraining and task-specific fine-tuning, we discover that some common practices are unnecessary with BERT models, such as using 
complex tagging schemes in named entity recognition (NER).
To help accelerate research in biomedical NLP, we have released our state-of-the-art pretrained and task-specific models for the community, and created a leaderboard featuring our BLURB benchmark (short for Biomedical Language Understanding \& Reasoning Benchmark) at \url{https://aka.ms/BLURB}.
\end{abstract}



\maketitle

\section{Introduction}
\label{sec:intro}

\begin{figure}
    \centering
    \includegraphics[width=0.9\textwidth]{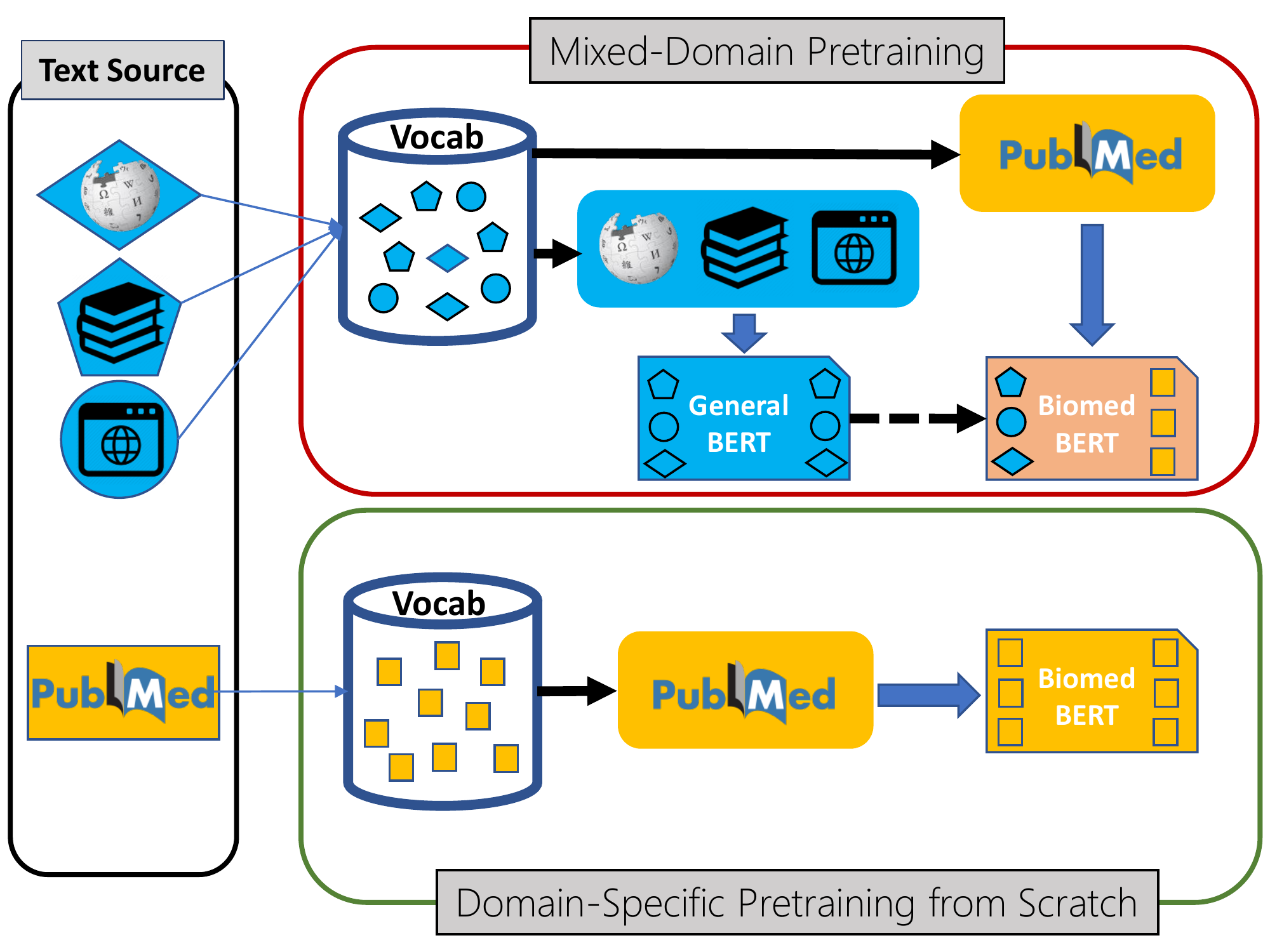}
    \caption{Two paradigms for neural language model pretraining. Top: The prevailing mixed-domain paradigm assumes that out-domain text is still helpful and typically initializes domain-specific pretraining with a general-domain language model and inherits its vocabulary. Bottom: Domain-specific pretraining from scratch derives the vocabulary and conducts pretraining using solely in-domain text. In this paper, we show that for domains with abundant text such as biomedicine, domain-specific pretraining from scratch can substantially outperform the conventional mixed-domain approach.
    }
    \label{fig:training_flow}
\end{figure}


In natural language processing (NLP), pretraining large neural language models on unlabeled text has proven to be a successful strategy for transfer learning. 
A prime example is Bidirectional Encoder Representations from Transformers~(BERT)~\cite{devlin2018bert}, which has become a standard building block for training task-specific NLP models.
Existing pretraining work typically focuses on the newswire and Web domains. For example, the original BERT model was trained on Wikipedia\footnote{\url{http://wikipedia.org}} and BookCorpus~\cite{zhu&al15_books}, and subsequent efforts have focused on crawling additional text from the Web to power even larger-scale pretraining~\cite{liu2019roberta,raffel2019t5}.

In specialized domains like biomedicine, past work has shown that using in-domain text can provide additional gains over general-domain language models~\cite{lee2019bioberts,beltagy2019scibert,peng2019transfer}.
However, a prevailing assumption is that out-domain text is still helpful and previous work typically adopts a mixed-domain approach, e.g., by starting domain-specific pretraining from an existing general-domain language model (\autoref{fig:training_flow} top).  
In this paper, we question this assumption. We observe that mixed-domain pretraining such as continual pretraining can be viewed as a form of transfer learning in itself, where the source domain is general text, such as newswire and the Web, and the target domain is specialized text such as biomedical papers.
Based on the rich literature of multi-task learning and transfer learning~\cite{axelrod-etal-2011-domain,xu-etal-2019-multi,caruana1997multitask,liu2015mtl}, successful transfer learning occurs when the target data is scarce and the source domain is highly relevant to the target one. 
For domains with abundant unlabeled text such as biomedicine, it is unclear that domain-specific pretraining can benefit by transfer from general domains. In fact, the majority of general domain text is substantively different from biomedical text, raising the prospect of negative transfer that actually hinders the target performance.


\eat{
To Pretrain or Not to Pretrain: Examining the Benefits of Pretraining on Resource Rich Tasks 
    this is presented at ACL 2020. 
main conclusion is: Pretraining might be reaching a diminishing return as the number of examples increases in downstream tasks; 

https://arxiv.org/pdf/2006.08671.pdf
another work is: (Nakkiran and Sutskever, 2020) observes that, in some translation task such as IWSLT14, small language models exhibit even lower test loss compared to the large transformer model when the number of training samples increases.
https://arxiv.org/pdf/1912.02292.pdf

---
how about this one? https://arxiv.org/pdf/1809.06963.pdf
here is another early paper: https://www.aclweb.org/anthology/D11-1033.pdf
The paper shows that during domain adaptation (transfer learning) instead of using all out-domain data, it is sometimes more efficient to use only a small subset of out-domain data (subcorpora) which is similar to the target domain.  "These subcorpora – 1
BTW, the first paper of above https://arxiv.org/pdf/1809.06963.pdf suggested that the out-domain data needs to be reweighted for transfer learning.
both papers support our claim "naive MTL doesn't work but needs instance reweighting or data selection".
}

We thus set out to conduct a rigorous study on domain-specific pretraining and its impact on downstream applications, using biomedicine as a running example.
{\em We show that domain-specific pretraining from scratch substantially outperforms continual pretraining of generic language models, thus demonstrating that the prevailing assumption in support of mixed-domain pretraining is not always applicable (\autoref{fig:training_flow}).}

To facilitate this study, we compile a comprehensive biomedical NLP benchmark from publicly-available datasets, and conduct in-depth comparisons of modeling choices for pretraining and task-specific fine-tuning by their impact on domain-specific applications.
Our experiments show that domain-specific pretraining from scratch can provide a solid foundation for biomedical NLP, leading to new state-of-the-art performance across a wide range of tasks.
Additionally, we discover that the use of transformer-based models, like BERT, necessitates rethinking several common practices. For example, BIO tags and more complex variants are the standard label representation for named entity recognition (NER). However, we find that simply using IO (in or out of entity mentions) suffices with BERT models, leading to comparable or better performance.

To help accelerate research in biomedical NLP, 
we have released our state-of-the-art pretrained and task-specific models for the community, and created a leaderboard featuring our comprehensive benchmark at \url{https://aka.ms/BLURB}.

\section{Methods}


\subsection{Language Model Pretraining}

In this section, we provide a brief overview of neural language model pretraining, using BERT~\cite{devlin2018bert} as a running example.

\subsubsection{Vocabulary}
\label{subsec:bert-vocab}
We assume that the input consists of text spans, such as sentences separated by special tokens~$\tt [SEP]$. To address the problem of out-of-vocabulary words, neural language models generate a vocabulary from subword units, using Byte-Pair Encoding~(BPE)~\cite{sennrich2015bpe} or variants such as WordPiece~\cite{kudo2018sentencepiece}.
Essentially, the BPE algorithm tries to greedily identify a small set of subwords that can compactly form all words in the given corpus. It does this by first shattering all words in the corpus and initializing the vocabulary with characters and delimiters.  It then iteratively augments the vocabulary with a new subword that is most frequent in the corpus and can be formed by concatenating two existing subwords, until the vocabulary reaches the pre-specified size (e.g., 30,000 in standard BERT models or 50,000 in RoBERTa~\cite{liu2019roberta}). In this paper, we use the WordPiece algorithm which is a BPE variant that uses likelihood based on the unigram language model rather than frequency in choosing which subwords to concatenate. 
The text corpus and vocabulary may preserve the case ($\tt cased$) or convert all characters to lower case ($\tt uncased$).


\subsubsection{Model Architecture}
\label{subsec:bert-mdl}
State-of-the-art neural language models are generally based on transformer architectures \cite{vaswani2017attention}, following the recent success of BERT ~\cite{devlin2018bert,liu2019roberta}.
The transformer model introduces a multi-layer, multi-head self-attention mechanism, which has demonstrated superiority in leveraging GPU-based parallel computation and modeling long-range dependencies in texts, compared to recurrent neural networks, such as LSTMs~\cite{hochreiter1997lstm}. 
The input token sequence is first processed by a lexical encoder, which combines a token embedding, a (token) position embedding and a segment embedding (i.e., which text span the token belongs to) by element-wise summation. 
This embedding layer is then passed to multiple layers of transformer modules \cite{vaswani2017attention}. 
In each transformer layer, a contextual representation is generated for each token by summing a non-linear transformation of the representations of all tokens in the prior layer, weighted by the attentions computed using the given token's representation in the prior layer as the query.
The final layer outputs contextual representations for all tokens, which combine information from the whole text span.



\subsubsection{Self-Supervision}
\label{subsec:bert-self-sup}
A key innovation in BERT \cite{devlin2018bert} is the use of a \textbf{Masked Language Model~(MLM)} for self-supervised pretraining. 
Traditional language models are typically generative models that predict the next token based on the preceding tokens; for example, n-gram models represent the conditional probability of the next token by a multinomial of the preceding n-gram, with various smoothing strategies to handle rare occurrences~\cite{kneser-ney}. 
Masked Language Model instead randomly replaces a subset of tokens by a special token (e.g., $\tt [MASK]$), and asks the language model to predict them. 
The training objective is the cross-entropy loss between the original tokens and the predicted ones. In BERT and RoBERTa, 15\% of the input tokens are chosen, among which a random 80\% are replaced by $\tt [MASK]$, 10\% are left unchanged and 10\% are randomly replaced by a token from the vocabulary. 
Instead of using a constant masking rate of 15\%, a standard approach is to gradually increase it from 5\% to 25\% with 5\% increment for every 20\% of training epochs, which makes pretraining more stable~\cite{liu2020alum}.  
The original BERT algorithm also uses \textbf{Next Sentence Prediction~(NSP)}, which determines for a given sentence pair whether one sentence follows the other in the original text. The utility of NSP has been called into question~\cite{liu2019roberta}, but we include it in our pretraining experiments to enable a head-to-head comparison with prior BERT models.

\subsubsection{Advanced Pretraining Techniques}
In the original formulation of BERT~\cite{devlin2018bert}, the masked language model (MLM) simply selects random subwords to mask. When a word is only partially masked, it is relatively easy to predict the masked portion given the observed ones. In contrast, whole-word masking (WWM) enforces that the whole word must be masked if one of its subwords is chosen. This has been adopted as the standard approach because it forces the language model to capture more contextual semantic dependencies. 

In this paper, we also explore adversarial pretraining and its impact on downstream applications. Motivated by successes in countering adversarial attacks in computer vision, adversarial pretraining introduces perturbations in the input embedding layer that maximize the adversarial loss, thus forcing the model to not only optimize the standard training objective (MLM), but also minimize adversarial loss~\cite{liu2020alum}. 

\subsection{Biomedical Language Model Pretraining}
\label{sec:domain}

In this paper, we will use biomedicine as a running example in our study of domain-specific pretraining. In other words, biomedical text is considered in-domain, while others are regarded as out-domain. 
Intuitively, using in-domain text in pretraining should help with domain-specific applications. Indeed, prior work has shown that pretraining with PubMed text leads to better performance in biomedical NLP tasks \cite{lee2019bioberts,beltagy2019scibert,peng2019transfer}. 
The main question is whether pretraining should include text from other domains. 
The prevailing assumption is that pretraining can always benefit from more text, including out-domain text. In fact, none of the prior biomedical-related BERT models have been pretrained using purely biomedical text \cite{lee2019bioberts,beltagy2019scibert,peng2019transfer}. 
Here, we challenge this assumption and show that {\em domain-specific pretraining from scratch} can be superior to {\em mixed-domain pretraining} for downstream applications.

\subsubsection{Mixed-Domain Pretraining}

The standard approach to pretraining a biomedical BERT model conducts {\em continual pretraining} of a general-domain pretrained model, as exemplified by BioBERT~\cite{lee2019bioberts}. 
Specifically, this approach would initialize with the standard BERT model~\cite{devlin2018bert}, pretrained using Wikipedia and BookCorpus. It then continues the pretraining process with MLM and NSP using biomedical text. 
In the case of BioBERT, continual pretraining is conducted using PubMed abstracts and PubMed Central full text articles. BlueBERT~\cite{peng2019transfer} uses both PubMed text and de-identified clinical notes from MIMIC-III~\cite{mimic}. 

Note that in the continual pretraining approach, the vocabulary is the same as the original BERT model, in this case the one generated from Wikipedia and BookCorpus. While convenient, this is a major disadvantage for this approach, as the vocabulary is not representative of the target biomedical domain.

Compared to the other biomedical-related pretraining efforts, SciBERT~\cite{beltagy2019scibert} is a notable exception as it generates the vocabulary and pretrains from scratch, using biomedicine and computer science as representatives for scientific literature. However, from the perspective of biomedical applications, SciBERT still adopts the mixed-domain pretraining approach, as computer science text is clearly out-domain.

\subsubsection{Domain-Specific Pretraining from Scratch}

The mixed-domain pretraining approach makes sense if the target application domain has little text of its own, and can thereby benefit from pretraining using related domains. 
However, this is not the case for biomedicine, which has over thirty million abstracts in PubMed, and adds over a million each year. 
{\em We thus hypothesize that domain-specific pretraining from scratch is a better strategy for biomedical language model pretraining.}

\eat{
\begin{table}[h]
    \centering
\begin{tabular}{llccc}
\specialrule{1pt}{1.5pt}{1.5pt}

              Biomedical Term &  Category & BERT & SciBERT & PubMedBERT (Ours) \\
\midrule
          diabetes &   disease &    \checkmark &       \checkmark &    \checkmark \\
          leukemia &   disease &    \checkmark &       \checkmark &    \checkmark \\
           lithium &  drug &    \checkmark &       \checkmark &    \checkmark \\
           insulin &      drug &    \checkmark &       \checkmark &    \checkmark \\
               DNA &      gene &    \checkmark &       \checkmark &    \checkmark \\
          promoter &      gene &    \checkmark &       \checkmark &    \checkmark \\
      hypertension &   disease &    [hyper, \#\#tension]  &       \checkmark &    \checkmark \\
       nephropathy &   disease &    [ne, \#\#ph, \#\#rop, \#\#athy]  &       \checkmark &    \checkmark \\
          lymphoma &   disease &    [l, \#\#ym, \#\#ph, \#\#oma]  &       \checkmark &    \checkmark \\
         lidocaine &  drug     &    [lid, \#\#oca, \#\#ine]  &       \checkmark &    \checkmark \\
     oropharyngeal &   organ   &    [oro, \#\#pha, \#\#ryn, \#\#ge, \#\#al]  &  [or, \#\#opharyngeal]       &    \checkmark \\
     cardiomyocyte &   cell    &    [card, \#\#iom, \#\#yo, \#\#cy, \#\#te]  &  [cardiomy, \#\#ocyte]       &    \checkmark \\
   chloramphenicol &      drug &    [ch, \#\#lor, \#\#amp, \#\#hen, \#\#ico, \#\#l]  &   [chlor, \#\#amp, \#\#hen, \#\#icol]      &    \checkmark \\
              RecA &      gene &    [rec, \#\#a]  &  [rec, \#\#a]       &    \checkmark \\
 acetyltransferase &      gene &    [ace, \#\#ty, \#\#lt, \#\#ran, \#\#sf, \#\#eras, \#\#e]  &   [acetyl, \#\#transferase]      &    \checkmark \\
         clonidine &  drug     &    [cl, \#\#oni, \#\#dine]   &   [clon, \#\#idine]      &    \checkmark \\
          naloxone &  drug     &    [na, \#\#lo, \#\#xon, \#\#e]  &  [nal, \#\#oxo, \#\#ne]       &    \checkmark \\
\bottomrule
\end{tabular}
    \caption{Comparison of common biomedical terms in vocabularies used by the standard BERT, SciBERT and PubMedBERT (ours). A~$\checkmark$~indicates the biomedical term appears in the corresponding vocabulary, otherwise the term will be shattered into small subwords as in $[ ]$. Although some shattered subwords are biomedical relevant terms, most of them do not preserve much meaning and thus probably hurt the model generalization to downstream tasks.}
    
    \label{tab:vocab_comparison}
\end{table}
} 
\begin{table}[h]
    \centering
\begin{tabular}{llccc}
\specialrule{1pt}{1.5pt}{1.5pt}

              Biomedical Term &  Category & BERT & SciBERT & PubMedBERT (Ours) \\
\midrule
          diabetes &   disease &    \checkmark &       \checkmark &    \checkmark \\
          leukemia &   disease &    \checkmark &       \checkmark &    \checkmark \\
           lithium &  drug &    \checkmark &       \checkmark &    \checkmark \\
           insulin &      drug &    \checkmark &       \checkmark &    \checkmark \\
               DNA &      gene &    \checkmark &       \checkmark &    \checkmark \\
          promoter &      gene &    \checkmark &       \checkmark &    \checkmark \\
      hypertension &   disease &    hyper-tension  &       \checkmark &    \checkmark \\
       nephropathy &   disease &    ne-ph-rop-athy  &       \checkmark &    \checkmark \\
          lymphoma &   disease &    l-ym-ph-oma  &       \checkmark &    \checkmark \\
         lidocaine &  drug     &    lid-oca-ine]  &       \checkmark &    \checkmark \\
     oropharyngeal &   organ   &    oro-pha-ryn-ge-al  &  or-opharyngeal       &    \checkmark \\
     cardiomyocyte &   cell    &    card-iom-yo-cy-te  &  cardiomy-ocyte       &    \checkmark \\
   chloramphenicol &      drug &    ch-lor-amp-hen-ico-l  &   chlor-amp-hen-icol      &    \checkmark \\
              RecA &      gene &    Rec-A  &  Rec-A       &    \checkmark \\
 acetyltransferase &      gene &    ace-ty-lt-ran-sf-eras-e  &   acetyl-transferase      &    \checkmark \\
         clonidine &  drug     &    cl-oni-dine   &   clon-idine      &    \checkmark \\
          naloxone &  drug     &    na-lo-xon-e  &  nal-oxo-ne       &   \checkmark \\
\bottomrule
\end{tabular}
    \caption{Comparison of common biomedical terms in vocabularies used by the standard BERT, SciBERT and PubMedBERT (ours). A~$\checkmark$~indicates the biomedical term appears in the corresponding vocabulary, otherwise the term will be broken into word pieces as separated by hyphen. These word pieces often have no biomedical relevance and may hinder learning in downstream tasks.
    }
    
    \label{tab:vocab_comparison}
\end{table}

A major advantage of domain-specific pretraining from scratch stems from having an in-domain vocabulary. 
\autoref{tab:vocab_comparison} compares the vocabularies used in various pretraining strategies. BERT models using continual pretraining are stuck with the original vocabulary from the general-domain corpora, which does not contain many common biomedical terms. 
Even for SciBERT, which generates its vocabulary partially from biomedical text, the deficiency compared to a purely biomedical vocabulary is substantial.
As a result, standard BERT models are forced to divert parametrization capacity and training bandwidth to model biomedical terms using fragmented subwords. 
For example, naloxone, a common medical term, is divided into four pieces ($[$na, \#\#lo, \#\#xon, \#\#e$]$) by BERT, and acetyltransferase is shattered into seven pieces ($[$ace, \#\#ty, \#\#lt, \#\#ran, \#\#sf, \#\#eras, \#\#e$]$) by BERT.\footnote{Prior work also observed similar shattering for clinical words \cite{si&al19}.} Both terms appear in the vocabulary of PubMedBERT.


%
%

Another advantage of domain-specific pretraining from scratch is that the language model is trained using purely in-domain data. 
For example, SciBERT pretraining has to balance optimizing for biomedical text and computer science text, the latter of which is unlikely to be beneficial for biomedical applications. 
Continual pretraining, on the other hand, may potentially recover from out-domain modeling, though not completely. Aside from the vocabulary issue mentioned earlier, neural network training uses non-convex optimization, which means that continual pretraining may not be able to completely undo suboptimal initialization from the general-domain language model. 

In our experiments, we show that domain-specific pretraining with in-domain vocabulary confers clear advantages over mixed-domain pretraining, be it continual pretraining of general-domain language models, or pretraining on mixed-domain text.




\subsection{BLURB: A Comprehensive Benchmark for Biomedical NLP}

\begin{table}[ht]
\begin{center}
\begin{tabular}{lcccc}
\specialrule{1pt}{1.5pt}{1.5pt}
&  BioBERT~\cite{lee2019bioberts} &  SciBERT~\cite{beltagy2019scibert} &  BLUE~\cite{peng2019transfer} &  BLURB \\
\midrule
BC5-chem~\cite{li2016biocreative}     &  \checkmark &  \checkmark &  \checkmark & \checkmark  \\
BC5-disease~\cite{li2016biocreative}     &  \checkmark &  \checkmark &  \checkmark & \checkmark  \\
NCBI-disease~\cite{dougan2014ncbi}     &  \checkmark &  \checkmark &  - & \checkmark  \\
BC2GM~\cite{smith2008overview}     &  \checkmark & - &  - & \checkmark  \\
JNLPBA~\cite{collier-kim-2004-introduction}     &  \checkmark &  - &  - & \checkmark  \\
\specialrule{0.05pt}{1.5pt}{1.5pt}
EBM PICO~\cite{nye2018corpus}     &  - &  \checkmark &  - & \checkmark  \\
\specialrule{0.05pt}{1.5pt}{1.5pt}
ChemProt~\cite{krallinger2017overview}     &  \checkmark &  \checkmark &  \checkmark & \checkmark  \\
DDI~\cite{herrero2013ddi}     &  \checkmark &  - &  \checkmark & \checkmark  \\
GAD~\cite{bravo2015extraction}     &  \checkmark &  - &  - & \checkmark  \\
\specialrule{0.05pt}{1.5pt}{1.5pt}
BIOSSES~\cite{souganciouglu2017biosses}     &  - &  - &  \checkmark & \checkmark  \\
\specialrule{0.05pt}{1.5pt}{1.5pt}
HoC~\cite{hanahan2000hallmarks}     &  - &  - & \checkmark & \checkmark  \\
\specialrule{0.05pt}{1.5pt}{1.5pt}
PubMedQA~\cite{jin2019PubMedqa}     & -  &  - &  - & \checkmark  \\
BioASQ~\cite{nentidis2019results}     &  \checkmark &  - &  - & \checkmark  \\
\specialrule{1pt}{1.5pt}{1.5pt}
\end{tabular}
\end{center}
\caption{Comparison of the biomedical datasets in prior language model pretraining studies and BLURB. \label{tab:paper-dataset}}
\end{table}



The ultimate goal of language model pretraining is to improve performance on a wide range of downstream applications. 
In general-domain NLP, the creation of comprehensive benchmarks, such as GLUE~\cite{wang19iclr_glue,wang2019superglue}, greatly accelerates advances in language model pretraining by enabling head-to-head comparisons among pretrained language models. 
In contrast, prior work on biomedical pretraining tends to use different tasks and datasets for downstream evaluation, as shown in \autoref{tab:paper-dataset}. This makes it hard to assess the impact of pretrained language models on the downstream tasks we care about. 
To the best of our knowledge, BLUE~\cite{peng2019transfer} is the first attempt to create an NLP benchmark in the biomedical domain. We aim to improve on its design by addressing some of its limitations. 
First, BLUE has limited coverage of biomedical applications used in other recent work on biomedical language models, as shown in \autoref{tab:paper-dataset}. For example, it does not include any question-answering task. More importantly, BLUE mixes PubMed-based biomedical applications (six datasets such as BC5, ChemProt, and HoC) with MIMIC-based clinical applications (four datasets such as i2b2 and MedNLI). Clinical notes differ substantially from biomedical literature, to the extent that we observe BERT models pretrained on clinical notes perform poorly on biomedical tasks, similar to the standard BERT. Consequently, it is advantageous to create separate benchmarks for these two domains.

To facilitate investigations of biomedical language model pretraining and help accelerate progress in biomedical NLP, we create a new benchmark, the {\em Biomedical Language Understanding \& Reasoning Benchmark (BLURB)}. 
We focus on PubMed-based biomedical applications, and leave the exploration of the clinical domain, and other high-value verticals to future work. 
To make our effort tractable and facilitate head-to-head comparison with prior work, we prioritize the selection of datasets used in recent work on biomedical language models, and will explore the addition of other datasets in future work. 


\begin{table}[ht]
\begin{center}
\begin{tabular}{llllll}
\specialrule{1pt}{1.5pt}{1.5pt}

Dataset  &  Task & Train & Dev & Test & Evaluation Metrics \\
\midrule
BC5-chem    &       NER &    5203 &         5347 &       5385 &        F1 entity-level \\
BC5-disease &       NER&    4182 &         4244 &       4424 &         F1 entity-level \\
NCBI-disease      &       NER&    5134 &     787     &     960   &          F1 entity-level \\
BC2GM      &       NER&     15197&         3061 &     6325   &          F1 entity-level \\
JNLPBA      &       NER&    46750&       4551   &    8662    &          F1 entity-level \\
\specialrule{0.05pt}{1.5pt}{1.5pt}
EBM PICO &       PICO&   339167 &         85321 &   16364     &         Macro F1 word-level \\
\specialrule{0.05pt}{1.5pt}{1.5pt}
ChemProt    &       Relation Extraction&    18035 &         11268 &       15745 &         Micro F1 \\
DDI         &       Relation Extraction&    25296 &         2496 &       5716 &        Micro F1 \\
GAD         &       Relation Extraction&    4261 &        535 &        534 &         Micro F1 \\
\specialrule{0.05pt}{1.5pt}{1.5pt}
BIOSSES     &       Sentence Similarity&    64&             16&       20 &         Pearson \\
\specialrule{0.05pt}{1.5pt}{1.5pt}
HoC         &       Document Classification&    1295 &         186 &       371 &          Micro F1\\
\specialrule{0.05pt}{1.5pt}{1.5pt}
PubMedQA    &       Question Answering& 450 &         50 &       500 &         Accuracy \\
BioASQ      &       Question Answering& 670 &         75 &       140 &         Accuracy \\
\bottomrule
\end{tabular}
\end{center}
\caption{Datasets used in the BLURB biomedical NLP benchmark. We list the numbers of instances in train, dev, and test (e.g., entity mentions in NER and PICO elements in evidence-based medical information extraction).
\label{tab:task-stats}
}
\end{table}

BLURB is comprised of a comprehensive set of biomedical NLP tasks from publicly available datasets, including named entity recognition (NER), 
evidence-based medical information extraction (PICO), relation extraction, sentence similarity, document classification, and question answering. 
See \autoref{tab:task-stats} for an overview of the BLURB datasets. 
For question answering, prior work has considered both classification tasks (e.g., whether a reference text contains the answer to a given question) and more complex tasks such as list and summary \cite{nentidis2019results}. The latter types often require additional engineering effort that are not relevant to evaluating neural language models. For simplicity, we focus on the classification tasks such as yes/no question-answering in BLURB, and leave the inclusion of more complex question-answering to future work.

To compute a summary score for BLURB, the simplest way is to report the average score among all tasks. However, this may place undue emphasis on simpler tasks such as NER for which there are many existing datasets. Therefore, we group the datasets by their task types, compute the average score for each task type, and report the macro average among the task types. 
To help accelerate research in biomedical NLP, we release the BLURB benchmark as well as a leaderboard at {\url{http://aka.ms/BLURB}}.

Below are detailed descriptions for each task and corresponding datasets.

\subsubsection{Named Entity Recognition (NER)}
\paragraph{BC5-Chemical \& BC5-Disease} The BioCreative V Chemical-Disease Relation corpus \cite{li2016biocreative} was created for evaluating relation extraction of drug-disease interactions, but is frequently used as a NER corpus for detecting chemical (drug) and disease entities. The dataset consists of 1500 PubMed abstracts broken into three even splits for training, development, and test. We use a pre-processed version of this dataset generated by \citet{crichton2017neural}, discard the relation labels, and train NER models for chemical (\textit{BC5-Chemical}) and disease (\textit{BC5-Disease}) separately.

\paragraph{NCBI-Disease} The Natural Center for Biotechnology Information Disease corpus \cite{dougan2014ncbi} contains 793 PubMed abstracts with 6892 annotated disease mentions linked to 790 distinct disease entities. We use a pre-processed set of train, development, test splits generated by \citet{crichton2017neural}.

\paragraph{BC2GM} The Biocreative II Gene Mention corpus \cite{smith2008overview} consists of sentences from PubMed abstracts with manually labeled gene and alternative gene entities. Following prior work, we focus on the gene entity annotation. In its original form, BC2GM contains 15000 train and 5000 test sentences. 
We use a pre-processed version of the dataset generated by \citet{crichton2017neural}, which carves out 2500 sentences from the training data for development.

\paragraph{JNLPBA} The Joint Workshop on Natural Language Processing in Biomedicine and its Applications shared task \cite{collier-kim-2004-introduction} is a NER corpus on PubMed abstracts. The entity types are chosen for molecular biology applications: protein, DNA, RNA, cell line, and cell type. 
Some of the entity type distinctions are not very meaningful. For example, a gene mention often refers to both the DNA and gene products such as the RNA and protein. 
Following prior work that evaluates on this dataset \cite{lee2019bioberts}, we ignore the type distinction and focus on detecting the entity mentions. 
We use the same train, development, and test splits as in \citet{crichton2017neural}.

\subsubsection{Evidence-Based Medical Information Extraction (PICO)}

\paragraph{EBM PICO} The Evidence-Based Medicine corpus \cite{nye2018corpus} contains PubMed abstracts on clinical trials, where each abstract is annotated with P, I, and O in PICO: Participants (e.g., \texttt{diabetic patients}), Intervention (e.g., \texttt{insulin}), Comparator (e.g., \texttt{placebo}) and Outcome (e.g., \texttt{blood glucose levels}). Comparator (C) labels are omitted as they are standard in clinical trials: placebo for passive control and standard of care for active control. There are 4300, 500, and 200 abstracts in training, development, and test, respectively. 
The training and development sets were labeled by Amazon Mechanical Turkers, whereas the test set was labeled by Upwork contributors with prior medical training. EBM PICO provides labels at the word level for each PIO element. For each of the PIO elements in an abstract, we tally the F1 score at the word level, and then compute the final score as the average among PIO elements in the dataset. Occasionally, two PICO elements might overlap with each other (e.g., a participant span might contain within it an intervention span). In EBM-PICO, about 3\% of the PIO words are in the overlap. Note that the dataset released along with SciBERT appears to remove the overlapping words from the larger span (e.g., the participant span as mentioned above). We instead use the original dataset~\cite{nye2018corpus} and their scripts for preprocessing and evaluation.

\subsubsection{Relation Extraction}
\paragraph{ChemProt} The Chemical Protein Interaction corpus \cite{krallinger2017overview} consists of PubMed abstracts annotated with chemical-protein interactions between chemical and protein entities. There are 23 interactions organized in a hierarchy, with 10 high-level interactions (including $\tt NONE$). 
The vast majority of relation instances in ChemProt are within single sentences. Following prior work \cite{lee2019bioberts,beltagy2019scibert}, we only consider sentence-level instances. 
We follow the ChemProt authors' suggestions and focus on classifying five high-level interactions --- $\tt UPREGULATOR~~(CPR:3)$, $\tt DOWNREGULATOR~~ (CPR:4)$, $\tt AGONIST~~ (CPR:5)$, $\tt ANTAGONIST~~ (CPR:6)$, $\tt SUBSTRATE ~~(CPR:9)$ --- as well as everything else ($\tt false$).
The ChemProt annotation is not exhaustive for all chemical-protein pairs.
Following previous work~\cite{peng2019transfer,lee2019bioberts}, we expand the training and development sets by assigning a $\tt false$ label for all chemical-protein pairs that occur in a training or development sentence, but do not have an explicit label in the ChemProt corpus. 
Note that prior work uses slightly different label expansion of the test data. To facilitate head-to-head comparison, we will provide instructions for reproducing the test set in BLURB from the original dataset.

\paragraph{DDI} The Drug-Drug Interaction corpus \cite{herrero2013ddi} was created to facilitate research on pharmaceutical information extraction, with a particular focus on pharmacovigilance. It contains sentence-level annotation of drug-drug interactions on PubMed abstracts. 
Note that some prior work \cite{peng2019transfer,zhang2018drug} discarded 90 training files that the authors considered not conducive to learning drug-drug interactions. We instead use the original dataset and produce our train/dev/test split of 624/90/191 files.

\paragraph{GAD} The Genetic Association Database corpus  \cite{bravo2015extraction} was created semi-automatically using the Genetic Association Archive.\footnote{\url{http://geneticassociationdb.nih.gov/}} Specifically, the archive contains a list of gene-disease associations, with the corresponding sentences in the PubMed abstracts reporting the association studies. Bravo et al.~\cite{bravo2015extraction} used a biomedical NER tool to identify gene and disease mentions, and create the positive examples from the annotated sentences in the archive, and negative examples from gene-disease co-occurrences that were not annotated in the archive. We use an existing preprocessed version of GAD and its corresponding train/dev/test split created by \citet{lee2019bioberts}.

\subsubsection{Sentence Similarity}
\paragraph{BIOSSES} The Sentence Similarity Estimation System for the Biomedical Domain \cite{souganciouglu2017biosses} contains 100 pairs of PubMed sentences each of which is annotated by five expert-level annotators with an estimated similarity score in the range from 0 (no relation) to 4 (equivalent meanings). It is a regression task, with the average score as the final annotation.
We use the same train/dev/test split in \citet{peng2019transfer} and use Pearson correlation for evaluation.

\subsubsection{Document Classification}

\paragraph{HoC} The Hallmarks of Cancer corpus was motivated by the pioneering work on cancer hallmarks \cite{hanahan2000hallmarks}. It contains annotation on PubMed abstracts with binary labels each of which signifies the discussion of a specific cancer hallmark. The authors use 37 fine-grained hallmarks which are grouped into ten top-level ones. We focus on predicting the top-level labels.
The dataset was released with 1499 PubMed abstracts \cite{baker2015automatic} and has since been expanded to 1852 abstracts \cite{baker2017cancer}. Note that \citet{peng2019transfer} discarded a control subset of 272 abstracts that do not discuss any cancer hallmark (i.e., all binary labels are false). 
We instead adopt the original dataset and report micro F1 across the ten cancer hallmarks. Though the original dataset provided sentence level annotation, we follow the common practice and evaluate on the abstract level \cite{6471714, 10.1093/jamia/ocz085}. We create the train/dev/test split, as they are not available previously.\footnote{The original authors used cross-validation for their evaluation.}

\subsubsection{Question Answering (QA)}

\paragraph{PubMedQA} The PubMedQA dataset \cite{jin2019PubMedqa} contains a set of research questions, each with a reference text from a PubMed abstract as well as an annotated label of whether the text contains the answer to the research question (yes/maybe/no). We use the original train/dev/test split with 450, 50, and 500 questions, respectively.

\paragraph{BioASQ} The BioASQ corpus \cite{nentidis2019results} contains multiple question answering tasks annotated by biomedical experts, including yes/no, factoid, list, and summary questions. 
Pertaining to our objective of comparing neural language models, we focus on the the yes/no questions (Task 7b), and leave the inclusion of other tasks to future work. Each question is paired with a reference text containing multiple sentences from a PubMed abstract and a yes/no answer. 
We use the official train/dev/test split of 670/75/140 questions.

\subsection{Task-Specific Fine-Tuning}
\label{sec:fine-tuning}

\begin{figure}
    \centering

	 \adjustbox{trim={0.1\width} {0.1\height} {0.1\width} {0.05\height},clip}{
     \includegraphics[width=1.0\textwidth]{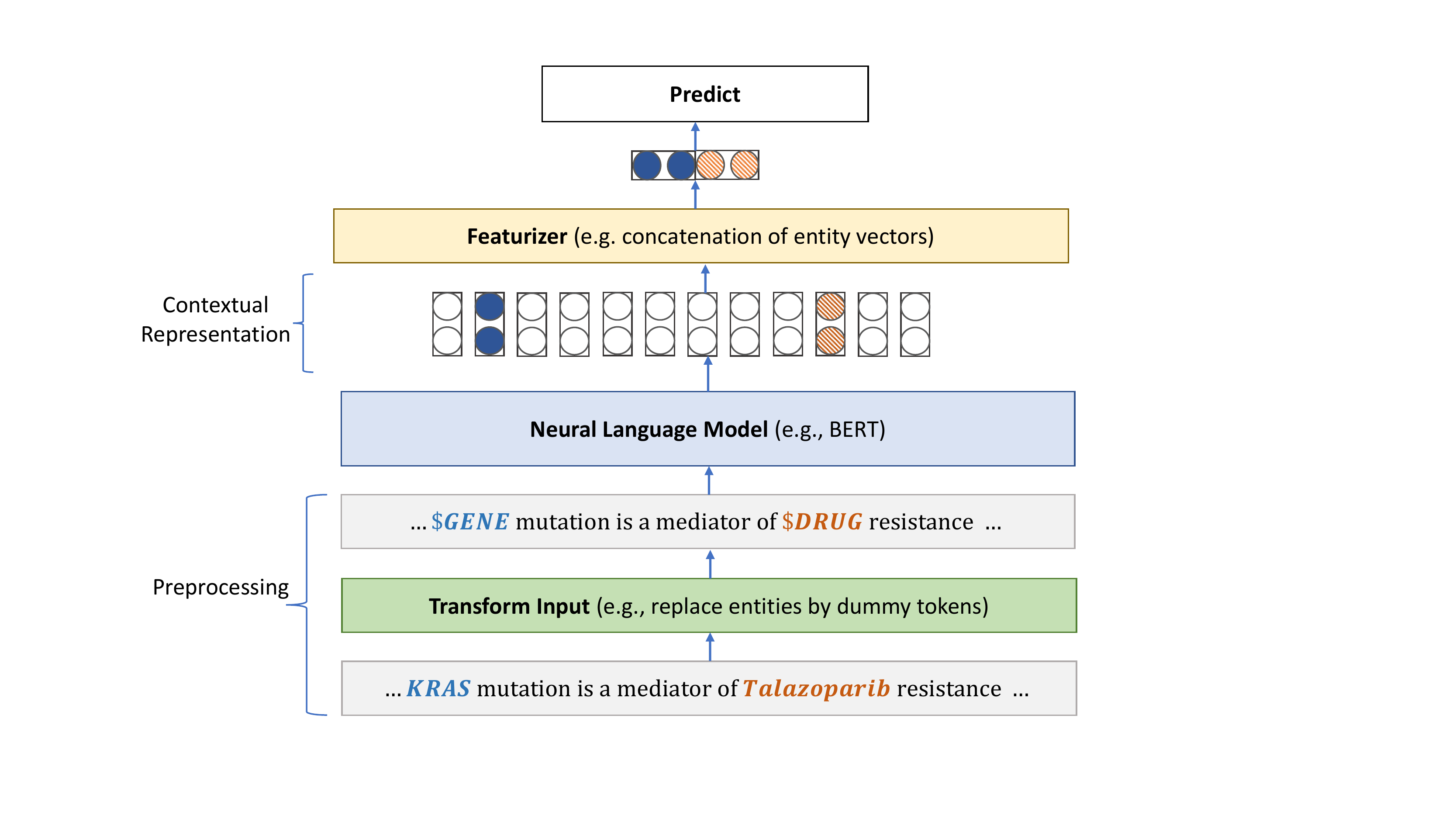}
     }
    \caption{A general architecture for task-specific fine-tuning of neural language models, with a relation-extraction example. Note that the input goes through additional processing such as word-piece tokenization in the neural language model module.
    }
    \label{fig:fine-tuning}
\end{figure}

\eat{
\begin{table}[h]
    \centering
\begin{tabular}{lll}
\toprule
            Task                    &  Problem Formulation        &  Epoch Range      \\
\specialrule{0.05pt}{1.5pt}{1.5pt}
            NER                         & Token Classification  & 20 - 60 \\
            PICO                        & Token Classification  & 2 - 3    \\
            Relation Extraction         & Sequence Classification   & 5 - 20 \\
            Sentence Similarity         & Sequence Regression       & 40 - 60 \\
            Document Classification     & Sequence Classification   & 10 - 30 \\
            Question Answering          & Sequence Classification   & 10 - 20 \\
\bottomrule
\end{tabular}
    \caption{Standard problem formulation}
    \label{tab:task_model_choices}
\end{table}

\begin{table}[h]
    \centering
\begin{tabular}{llll}
\toprule
                 Model &         Components          &  Standard     &   Variants \\
\specialrule{0.05pt}{1.5pt}{1.5pt}
            Sequence Classification/Regression &    Hidden States Output    & [CLS]         &   Entity Tokens \\
            Sequence Classification/Regression &    Pooling                 &  -            &   Max Pool.; Mean Pool.; Entity Start\\
            Sequence Classification/Regression &    Top Layer               & Linear Layer  &   Bi-LSTM  \\
            Token Classification & Tagging                 & BIO           &   BIOUL; IO \\
            Token Classification& Top Layer               & Linear Layer  &   Bi-LSTM  \\

\bottomrule
\end{tabular}
    \caption{Model Variants.}
    \label{tab:model_var}
\end{table}
}

Pretrained neural language models provide a unifying foundation for learning task-specific models. 
Given an input token sequence, the language model produces a sequence of vectors in the contextual representation. A task-specific prediction model is then layered on top to generate the final output for a task-specific application.
Given task-specific training data, we can learn the task-specific model parameters and refine the BERT model parameters by gradient descent using backpropragation.

Prior work on biomedical NLP often adopts different task-specific models and fine-tuning methods, which makes it difficult to understand the impact of an underlying pretrained language model on task performance. 
In this section, we review standard methods and common variants used for each task.
In our primary investigation comparing pretraining strategies, we fix the task-specific model architecture using the standard method identifed here, to facilitate a head-to-head comparison among the pretrained neural language models. 
Subsequently, we start with the same pretrained BERT model, and conduct additional investigation on the impact for the various choices in the task-specific models.
For prior biomedical BERT models, our standard task-specific methods generally lead to comparable or better performance when compared to their published results.

\subsubsection{A General Architecture for Fine-Tuning Neural Language Models}

\autoref{fig:fine-tuning} shows a general architecture of fine-tuning neural language models for downstream applications. 
An input instance is first processed by a $\tt TransformInput$ module which performs task-specific transformations such as appending special instance marker (e.g., $\tt [CLS]$) or dummifying entity mentions for relation extraction.
The transformed input is then tokenized using the neural language model's vocabulary, and fed into the neural language model. Next, the contextual representation at the top layer is processed by a $\tt Featurizer$ module, and then fed into the $\tt Predict$ module to generate the final output for a given task. 

To facilitate a head-to-head comparison, we apply the same fine-tuning procedure for all BERT models and tasks. Specifically, we use cross-entropy loss for classification tasks and mean-square error for regression tasks. We conduct hyperparameter search using the development set based on task-specific metrics. Similar to previous work, we jointly fine-tune the parameters of the task-specific prediction layer as well as the underlying neural language model.

\begin{table}[h]
    \centering
\begin{tabular}{lll}
\specialrule{1pt}{1.5pt}{1.5pt}

            Task                    &  Problem Formulation     & Modeling Choices\\
\midrule
            NER                         & Token Classification  & Tagging Scheme, Classification Layer\\
            PICO                        & Token Classification  & Tagging Scheme, Classification Layer\\
            Relation Extraction         & Sequence Classification   & Entity/Relation Representation, Classification Layer\\
            Sentence Similarity         & Sequence Regression       & Sentence Representation, Regression Loss\\
            Document Classification     & Sequence Classification   & Document Representation, Classification Layer\\
            Question Answering          & Sequence Classification   & Question/Text Representation, Classification Layer\\
\specialrule{1pt}{1.5pt}{1.5pt}
\end{tabular}
    \caption{Standard NLP tasks and their problem formulations and modeling choices.}
    \label{tab:task_model_choices}
\end{table}

\subsubsection{Task-Specific Problem Formulation and Modeling Choices}
Many NLP applications can be formulated as a classification or regression task, wherein either individual tokens or sequences are the prediction target. Modeling choices usually vary in two aspects: the instance representation and the prediction layer. \autoref{tab:task_model_choices} presents an overview of the problem formulation and modeling choices for tasks we consider and detailed descriptions are provided below. For each task, we highlight the standard modeling choices with an asterisk~(*).

\paragraph{NER} Given an input text span (usually a sentence), the NER task seeks to recognize mentions of entities of interest. It is typically formulated as a sequential labeling task, where each token is assigned a tag to signify whether it is in an entity mention or not. The modeling choices primarily vary on the tagging scheme and classification method. $\tt BIO$ is the standard tagging scheme that classifies each token as the beginning of an entity~($\tt B$), inside an entity~($\tt I$), or outside~($\tt O$). The NER tasks in BLURB are only concerned about one entity type (in JNLPBA, all the types are merged into one). In the case when there are multiple entity types, the $\tt BI$ tags would be further divided into fine-grained tags for specific types. Prior work has also considered more complex tagging schemes such as $\tt BIOUL$, where $\tt U$ stands for the last word of an entity and $\tt L$ stands for a single-word entity. We also consider the simpler $\tt IO$ scheme that only differentiates between in and out of an entity. Classification is done using a simple linear layer or more sophisticated sequential labeling methods such as LSTM or conditional random field~(CRF)~\cite{Lafferty01conditionalrandom}.

\begin{itemize}
    \item $\tt TransformInput$: returns the input sequence as is.
    \item $\tt Featurizer$: returns the BERT encoding of a given token.
    \item Tagging scheme: $\tt BIO$*; $\tt BIOUL$; $\tt IO$.
    \item Classification layer: linear layer*; LSTM; CRF.
\end{itemize}

\paragraph{PICO} Conceptually, evidence-based medical information extraction is akin to slot filling, as it tries to identify the PIO elements in an abstract describing a clinical trial. However, it can be formulated as a sequential tagging task like NER, by classifying tokens belonging to each element. A token may belong to more than one element, e.g., participant (P) and intervention (I).  
\begin{itemize}
    \item $\tt TransformInput$: returns the input sequence as is.
    \item $\tt Featurizer$: returns the BERT encoding of a given token.
    \item Tagging scheme: $\tt BIO$*; $\tt BIOUL$; $\tt IO$.
    \item Classification layer: linear layer*; LSTM; CRF.
\end{itemize}

\paragraph{Relation Extraction} Existing work on relation extraction tends to focus on binary relations. Given a pair of entity mentions in a text span (typically a sentence), the goal is to determine if the text indicates a relation for the mention pair. 
There are significant variations in the entity and relation representations. To prevent overfitting by memorizing the entity pairs, the entity tokens are often augmented with start/end markers or replaced by a dummy token. For featurization, the relation instance is either represented by a special $\tt [CLS]$ token, or by concatenating the mention representations. In the latter case, if an entity mention contains multiple tokens, its representation is usually produced by pooling those of individual tokens (max or average). For computational efficiency, we use padding or truncation to set the input length to 128 tokens for GAD and 256 tokens for ChemProt and DDI which contain longer input sequences.
\begin{itemize}
    \item $\tt TransformInput$: entity (dummification*; start/end marker; original); relation ($\tt [CLS]$*; original).
    \item $\tt Featurizer$: entity (dummy token*; pooling); relation ($\tt [CLS]$ BERT encoding*; concatenation of the mention BERT encoding).
    \item Classification layer: linear layer*; more sophisticated classifiers (e.g., MLP).
\end{itemize}

\paragraph{Sentence Similarity} The similarity task can be formulated as a regression problem to generate a normalized score for a sentence pair. By default, a special $\tt [SEP]$ token is inserted to separate the two sentences, and a special $\tt [CLS]$ token is prepended to the beginning to represent the pair. The BERT encoding of $\tt [CLS]$ is used to compute the regression score.
\begin{itemize}
    \item $\tt TransformInput$: $\tt [CLS]$ $S_1$ $\tt [SEP]$ $S_2$ $\tt [SEP]$, for sentence pair $S_1, S_2$.
    \item $\tt Featurizer$: $\tt [CLS]$ BERT encoding.
    \item Regression layer: linear regression.
\end{itemize}

\paragraph{Document Classification} For each text span and category (an abstract and a cancer hallmark in HoC), the goal is to classify whether the text belongs to the  category. By default, a $\tt [CLS]$ token is appended to the beginning of the text, and its BERT encoding is passed on by the $\tt Featurizer$ for the final classification, which typically uses a simple linear layer.
\begin{itemize}
    \item $\tt TransformInput$: $\tt [CLS]$ $D$ $\tt [SEP]$, for document $D$.
    \item $\tt Featurizer$: returns $\tt [CLS]$ BERT encoding.
    \item Classification layer: linear layer.
\end{itemize}


\paragraph{Question Answering} For the two-way (yes/no) or three-way (yes/maybe/no) question-answering task, the encoding is similar to the sentence similarity task. Namely, a $\tt [CLS]$ token is prepended to the beginning, followed by the question and reference text, with a $\tt [SEP]$ token to separate the two text spans. The $\tt [CLS]$ BERT encoding is then used for the final classification. For computational efficiency, we use padding or truncation to set the input length to 512 tokens.
\begin{itemize}
    \item $\tt TransformInput$: $\tt [CLS]$ $Q$ $\tt [SEP]$ $T$ $\tt [SEP]$, for question $Q$ and reference text $T$.
    \item $\tt Featurizer$: returns $\tt [CLS]$ BERT encoding. 
    \item Classification layer: linear layer.
\end{itemize}

\subsection{Experimental Settings}

\begin{table}[ht]
\begin{center}
\begin{tabular}{lccccc}
\specialrule{1pt}{1.5pt}{1.5pt}
&  Vocabulary & Pretraining & Corpus & Text Size \\
\midrule
BERT & Wiki + Books & - & Wiki + Books & 3.3B words / 16GB \\
RoBERTa & Web crawl & - & Web crawl & 160GB \\
BioBERT & Wiki + Books & continual pretraining & PubMed & 4.5B words \\
SciBERT & PMC + CS & from scratch & PMC + CS & 3.2B words \\
ClinicalBERT & Wiki + Books & continual pretraining & MIMIC & 0.5B words / 3.7GB \\
BlueBERT & Wiki + Books & continual pretraining & PubMed + MIMIC & 4.5B words \\
PubMedBERT & PubMed & from scratch & PubMed & 3.1B words / 21GB \\
\specialrule{1pt}{1.5pt}{1.5pt}
\end{tabular}
\end{center}
\caption{Summary of pretraining details for the various BERT models used in our experiments. Statistics for prior BERT models are taken from their publications when available. The size of a text corpus such as PubMed may vary a bit, depending on downloading time and preprocessing (e.g., filtering out empty or very short abstracts). Both BioBERT and PubMedBERT also have a version pretrained with additional PMC full text; here we list the standard version pretrained using PubMed only.}
\label{tab:bert-pretraining}
\end{table}

For biomedical domain-specific pretraining, we generate the vocabulary and conduct pretraining using the latest collection of PubMed\footnote{\url{https://pubmed.ncbi.nlm.nih.gov/}; downloaded in Feb. 2020.} abstracts: 14 million abstracts, 3.2 billion words, 21 GB. (The original collection contains over 4 billion words; we filter out any abstracts with less than 128 words to reduce noise.)

We follow the standard pretraining procedure based on the Tensorflow implementation released by NVIDIA.\footnote{\url{https://github.com/NVIDIA/DeepLearningExamples}} 
We use Adam~\cite{kingma2014adam} for the optimizer using a standard slanted triangular learning rate schedule with warm-up in 10\% of steps and cool-down in 90\% of steps. Specifically, the learning rate increases linearly from zero to the peak rate of $6\times 10^{-4}$ in the first 10\% of steps, and then decays linearly to zero in the remaining 90\% of steps. 
Training is done for 62,500 steps with batch size of 8,192, which is comparable to the computation used in previous biomedical pretraining.\footnote{For example, BioBERT started with the standard BERT, which was pretrained using 1M steps with batch size of 256, and ran another 1M steps in continual pretraining.}
The training takes about 5 days on one DGX-2 machine with 16 V100 GPUs. 
We find that the cased version has similar performance to the uncased version in preliminary experiments; thus, we focus on uncased models in this study.
We use whole-word masking (WWM), with a masking rate of 15\%. We denote the resulting BERT model {\em PubMedBERT}.

For comparison, we use the public releases of {BERT}~\cite{devlin2018bert}, {RoBERTa}~\cite{liu2019roberta}, {BioBERT}~\cite{lee2019bioberts}, {SciBERT}~\cite{beltagy2019scibert}, {ClinicalBERT}~\cite{alsentzer-etal-2019-publicly}, and {BlueBERT}~\cite{peng2019transfer}. 
See \autoref{tab:bert-pretraining} for an overview. 
BioBERT and BlueBERT conduct continual pretraining from BERT, whereas ClinicalBERT conducts continual pretraining from BioBERT; thus, they all share the same vocabulary as BERT. 
BioBERT comes with two versions. We use BioBERT++ (v1.1), which was trained for a longer time and performed better. ClinicalBERT also comes with two versions. We use Bio+Clinical BERT.

Prior pretraining work has explored two settings: BERT-BASE with 12 transformer layers and 100 million parameters; BERT-LARGE with 24 transformer layers and 300 million parameters. Prior work in biomedical pretraining uses BERT-BASE only. For head-to-head comparison, we also use BERT-BASE in pretraining PubMedBERT. BERT-LARGE appears to yield improved performance in some preliminary experiments. We leave an in-depth exploration to future work. 


For task-specific fine-tuning, we use Adam~\cite{kingma2014adam} with the standard slanted triangular learning rate schedule (warm-up in the first 10\% of steps and cool-down in the remaining 90\% of steps) and a dropout probability of 0.1. 
Due to random initialization of the task-specific model and drop out, the performance may vary for different random seeds, especially for small datasets like BIOSSES, BioASQ, and PubMedQA. We report the average scores from ten runs for BIOSSES, BioASQ, and PubMedQA, and five runs for the others. 

For all datasets, we use the development set for tuning the hyperparameters with the same range: learning rate (1e-5, 3e-5, 5e-5), batch size (16, 32) and epoch number (2--60). 
Ideally, we would conduct separate hyperparameter tuning for each model on each dataset. However, this would incur a prohibitive amount of computation, as we have to enumerate all combinations of models, datasets and hyperparameters, each of which requires averaging over multiple runs with different randomization. In practice, we observe that the development performance is not very sensitive to hyperparameter selection, as long as they are in a ballpark range. Consequently, we focus on hyperparameter tuning using a subset of representative models such as BERT and BioBERT, and use a common set of hyperparameters for each dataset that work well for both out-domain and in-domain language models.

\section{Results}
\label{sec:results}

In this section, we conduct a thorough evaluation to assess the impact of domain-specific pretraining in biomedical NLP applications. First, we fix the standard task-specific model for each task in BLURB, and conduct a head-to-head comparison of domain-specific pretraining and mixed-domain pretraining.
Next, we 
evaluate the impact of various pretraining options such as vocabulary, whole-word masking (WWM), and adversarial pretraining.
Finally, we fix a pretrained BERT model and compare various modeling choices for task-specific fine-tuning.

\subsection{Domain-Specific Pretraining vs Mixed-Domain Pretraining}




\begin{table}[ht]
\begin{center}
\resizebox{\textwidth}{!}{\begin{tabular}{lccccccccc}
\specialrule{1pt}{1.5pt}{1.5pt}
&  \multicolumn{2}{c}{BERT} &  RoBERTa & BioBERT & \multicolumn{2}{c}{SciBERT} & ClinicalBERT & BlueBERT &  PubMedBERT \\
&    uncased &  cased &    cased &   cased &       uncased &  cased &  cased &    cased &     uncased \\
\midrule
BC5-chem     &      89.25 &  89.99 &    89.43 &    92.85 &    92.49 &  92.51 &         90.80 &     91.19 &       \textbf{93.33} \\
BC5-disease  &      81.44 &  79.92 &    80.65 &    84.70 &    84.54 &  84.70 &         83.04 &     83.69 &       \textbf{85.62} \\
NCBI-disease &      85.67 &  85.87 &    86.62 &    \textbf{89.13} &    88.10 &  88.25 &         86.32 &     88.04 &       87.82 \\
BC2GM        &      80.90 &  81.23 &    80.90 &    83.82 &    83.36 &  83.36 &         81.71 &     81.87 &       \textbf{84.52} \\
JNLPBA       &      77.69 &  77.51 &    77.86 &    78.55 &    78.68 &  78.51 &         78.07 &     77.71 &       \textbf{79.10} \\
\specialrule{0.05pt}{1.5pt}{1.5pt}
EBM PICO     &      72.34 &  71.70 &    73.02 &    73.18 &    73.12 &  73.06 &         72.06 &     72.54 &       \textbf{73.38} \\
\specialrule{0.05pt}{1.5pt}{1.5pt}
ChemProt     &      71.86 &  71.54 &    72.98 &    76.14 &    75.24 &  75.00 &         72.04 &     71.46 &       \textbf{77.24} \\
DDI          &      80.04 &  79.34 &    79.52 &    80.88 &    81.06 &  81.22 &         78.20 &     77.78 &       \textbf{82.36} \\
GAD          &      80.41 &  79.61 &    80.63 &    82.36 &    82.38 &  81.34 &         80.48 &     79.15 &       \textbf{83.96} \\
BIOSSES      &      82.68 &  81.40 &    81.25 &    89.52 &    86.25 &  87.15 &         91.23 &     85.38 &       \textbf{92.30} \\
\specialrule{0.05pt}{1.5pt}{1.5pt}
HoC          &      80.20 &  80.12 &    79.66 &    81.54 &    80.66 &  81.16 &         80.74 &     80.48 &       \textbf{82.32} \\
\specialrule{0.05pt}{1.5pt}{1.5pt}
PubMedQA     &      51.62 &  49.96 &    52.84 &    \textbf{60.24} &    57.38 &  51.40 &         49.08 &     48.44 &       55.84 \\
BioASQ       &      70.36 &  74.44 &    75.20 &    84.14 &    78.86 &  74.22 &         68.50 &     68.71 &       \textbf{87.56} \\
\specialrule{0.05pt}{1.5pt}{1.5pt}
BLURB score  &      76.11 &  75.86 &    76.46 &    80.34 &    78.86 &  78.14 &         77.29 &     76.27 &       \textbf{81.16} \\
\specialrule{1pt}{1.5pt}{1.5pt}
\end{tabular}}
\end{center}
\caption{Comparison of pretrained language models on the BLURB biomedical NLP benchmark. The standard task-specific models are used in the same fine-tuning process for all BERT models. The BLURB score is the macro average of average test results for each of the six tasks (NER, PICO, relation extraction, sentence similarity, document classification, question answering). See \autoref{tab:task-stats} for the evaluation metric used in each task.}
\label{tab:bert-compare}
\end{table}


We compare BERT models by applying them to the downstream NLP applications in BLURB.
For each task, we conduct the same fine-tuning process using the standard task-specific model as specified in \autoref{sec:fine-tuning}. %
\autoref{tab:bert-compare} shows the results. 

By conducting domain-specific pretraining from scratch, PubMedBERT consistently outperforms all the other BERT models in most biomedical NLP tasks, often by a significant margin.
The gains are most substantial against BERT models trained using out-domain text. 
Notably, while the pretraining corpus is the largest for RoBERTa, its performance on biomedical NLP tasks is among the worst, similar to the original BERT model. 
Models using biomedical text in pretraining generally perform better. However, mixing out-domain data in pretraining generally leads to worse performance. In particular, even though clinical notes are more relevant to the biomedical domain than general-domain text, adding them does not confer any advantage, as evident by the results of ClinicalBERT and BlueBERT. 
Not surprisingly, BioBERT is the closest to PubMedBERT, as it also uses PubMed text for pretraining. However, by conducting domain-specific pretraining from scratch, including using the PubMed vocabulary, PubMedBERT is able to obtain consistent gains over BioBERT in most tasks. A notable exception is PubMedQA, but this dataset is small, and there are relatively high variances among runs with different random seeds.

Compared to the published results for BioBERT, SciBERT, and BlueBERT in their original papers, our results are generally comparable or better for the tasks they have been evaluated on. The ClinicalBERT paper does not report any results on these biomedical applications \cite{alsentzer-etal-2019-publicly}.

\eat{
\begin{table}[ht]
\begin{center}
\begin{tabular}{lcccc}
\specialrule{1pt}{1.5pt}{1.5pt}
&  BioBERT paper &  SciBERT paper &  BLUE paper &  BLURB \\
\midrule
BC5-chem     &  \checkmark &  \checkmark &  \checkmark & \checkmark  \\
BC5-disease     &  \checkmark &  \checkmark &  \checkmark & \checkmark  \\
NCBI-disease     &  \checkmark &  \checkmark &  - & \checkmark  \\
BC2GM     &  \checkmark & - &  - & \checkmark  \\
JNLPBA     &  \checkmark &  - &  \checkmark & \checkmark  \\
\specialrule{0.05pt}{1.5pt}{1.5pt}
EBM PICO     &  - &  \checkmark &  - & \checkmark  \\
\specialrule{0.05pt}{1.5pt}{1.5pt}
ChemProt     &  \checkmark &  \checkmark &  \checkmark & \checkmark  \\
DDI     &  \checkmark &  - &  \checkmark & \checkmark  \\
GAD     &  \checkmark &  - &  - & \checkmark  \\
\specialrule{0.05pt}{1.5pt}{1.5pt}
BIOSSES     &  - &  - &  \checkmark & \checkmark  \\
\specialrule{0.05pt}{1.5pt}{1.5pt}
HoC     &  - &  - & \checkmark & \checkmark  \\
\specialrule{0.05pt}{1.5pt}{1.5pt}
PubMedQA     & -  &  - &  - & \checkmark  \\
BioASQ     &  \checkmark &  - &  - & \checkmark  \\
\specialrule{1pt}{1.5pt}{1.5pt}
\end{tabular}
\end{center}
\caption{Comparison of the biomedical datasets in prior language model pretraining studies and BLURB.}
\label{tab:paper-dataset}
\end{table}

Prior work on biomedical pretraining tends to use different tasks and datasets for evaluation, as we can see in \autoref{tab:paper-dataset}. This makes it harder to assess the true impact of various BERT models in biomedical applications. By creating BLURB, we hope to fill in this gap and facilitate systematic comparison of language model pretraining for biomedical NLP. As we can see above, the results clearly elucidate the impact of domain-specific pretraining compared to mixed-domain pretraining.
}

\subsection{Ablation Study on Pretraining Techniques}

\begin{table}[ht]
\begin{center}
\begin{tabular}{lcccc}
\specialrule{1pt}{1.5pt}{1.5pt}
      &  \multicolumn{2}{c}{Wiki + Books} &  \multicolumn{2}{c}{PubMed} \\
   &    Word Piece &  Whole Word &   Word Piece &  Whole Word \\
\midrule
BC5-chem     &   93.20 &  93.31 &   92.96 &   \textbf{93.33} \\
BC5-disease  &   85.00 &  85.28 &   84.72 &   \textbf{85.62} \\
NCBI-disease &   88.39 &  \textbf{88.53} &   87.26 &   87.82 \\
BC2GM        &   83.65 &  83.93 &   83.19 &   \textbf{84.52} \\
JNLPBA       &   78.83 &  78.77 &   78.63 &   \textbf{79.10} \\
\specialrule{0.05pt}{1.5pt}{1.5pt}
EBM PICO     &   73.30 &  \textbf{73.52} &   73.44 &   73.38 \\
\specialrule{0.05pt}{1.5pt}{1.5pt}
ChemProt     &   75.04 &  76.70 &   75.72 &   \textbf{77.24} \\
DDI          &   81.30 &  \textbf{82.60} &   80.84 &   82.36 \\
GAD          &   83.02 &  82.42 &   81.74 &   \textbf{83.96} \\
\specialrule{0.05pt}{1.5pt}{1.5pt}
BIOSSES      &   91.36 &  91.79 &   \textbf{92.45} &   92.30 \\
\specialrule{0.05pt}{1.5pt}{1.5pt}
HoC          &   81.76 &  81.74 &   80.38 &   \textbf{82.32} \\
\specialrule{0.05pt}{1.5pt}{1.5pt}
PubMedQA     &   52.20 &  \textbf{55.92} &   54.76 &   55.84 \\
BioASQ       &   73.69 &  76.41 &   78.51 &   \textbf{87.56} \\
\specialrule{0.05pt}{1.5pt}{1.5pt}
BLURB score  &   79.16 &  79.96 &   79.62 &   \textbf{81.16} \\
\bottomrule
\end{tabular}
\end{center}
\caption{Evaluation of the impact of vocabulary and whole word masking on the performance of PubMedBERT on BLURB.}
\label{tab:ablation-vocab-wwm}
\end{table}

\begin{figure}
    \centering
    \begin{subfigure}{0.9\linewidth}
    \includegraphics[width=\linewidth]{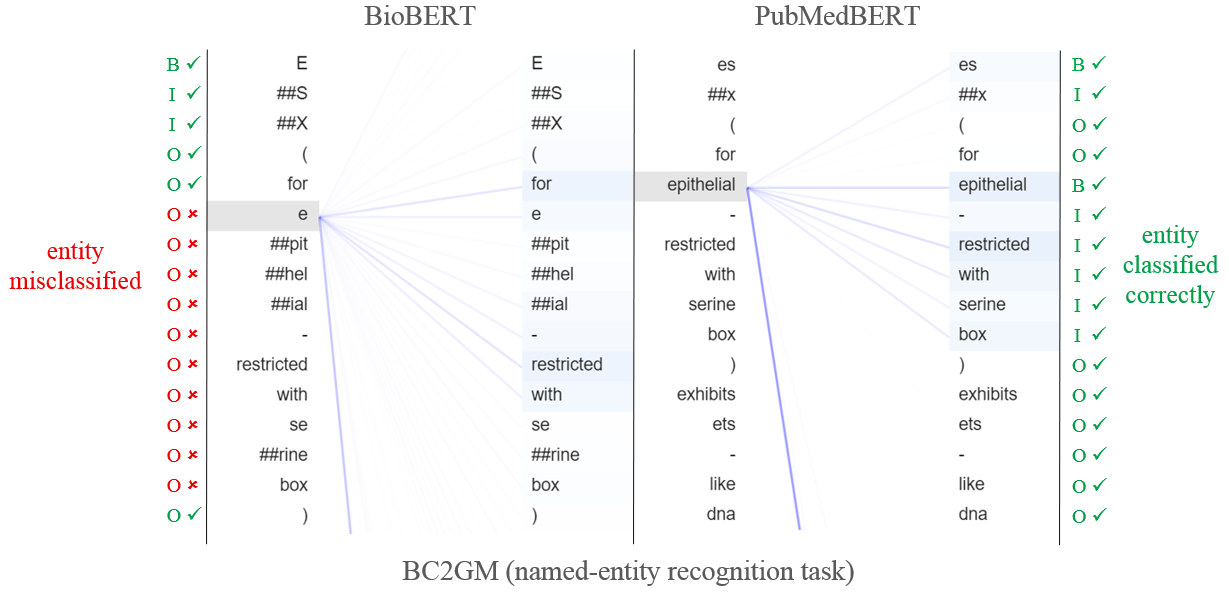}
    \end{subfigure}
    \newline
    \begin{subfigure}{0.8\linewidth}
    \includegraphics[width=\linewidth]{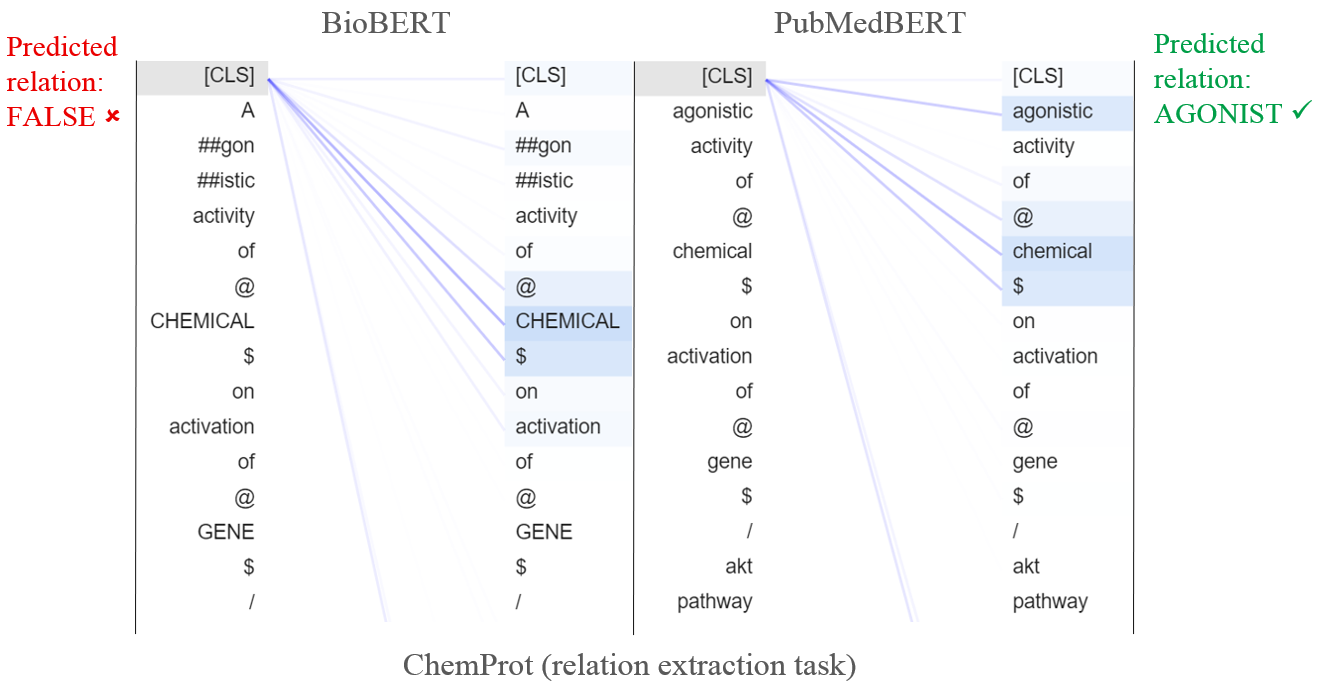}
    \end{subfigure}
    \caption{
    Examples of how domain-specific pretraining helps correct errors from mixed-domain pretraining. Top: attention for the leading word piece of the gene mention ``epithelial-restricted with serine box" (abbreviation ``ESX") in the BC2GM dataset.
    Bottom: attention for the [CLS] token in an instance of AGONIST relation between a pair of dummified chemical and protein. 
    In both cases, we show the aggregate attention from the penultimate layer to the preceding layer, which tends to be most informative about the final classification. Note how BioBERT tends to shatter the relevant words by inheriting the general-domain vocabulary. The domain-specific vocabulary enables PubMedBERT to learn better attention patterns and make correct predictions.
    }
    \label{fig:attention-examples}
\end{figure}

\begin{table}[ht]
\begin{center}
\begin{tabular}{lcc}
\specialrule{1pt}{1.5pt}{1.5pt}
Vocab & Wiki + Books & PubMed \\
\midrule
BC5-chem     &   35.9 & 28.0 \\
BC5-disease  &   35.9 & 28.0 \\
NCBI-disease &   34.2 & 27.4 \\
BC2GM        &   38.5 & 30.5 \\
JNLPBA       &   33.7 & 26.0 \\
\specialrule{0.05pt}{1.5pt}{1.5pt}
EBM PICO     &   30.7 & 25.1 \\
\specialrule{0.05pt}{1.5pt}{1.5pt}
ChemProt     &   75.4 & 55.5 \\
DDI          &   106.0 & 75.9 \\
GAD          &   47.0 & 35.7 \\
\specialrule{0.05pt}{1.5pt}{1.5pt}
BIOSSES      &   80.7 & 61.6 \\
\specialrule{0.05pt}{1.5pt}{1.5pt}
HoC          &   40.6 & 31.0 \\
\specialrule{0.05pt}{1.5pt}{1.5pt}
PubMedQA     &   343.1 & 293.0 \\
BioASQ       &   702.4 & 541.4 \\
\bottomrule
\end{tabular}
\end{center}
\caption{Comparison of the average input length in word pieces using  general-domain vs in-domain vocabulary.}
\label{tab:input-length}
\end{table}

\begin{table}[ht]
\begin{center}
\begin{tabular}{lcccc}
\specialrule{1pt}{1.5pt}{1.5pt}
Pretraining &  \multicolumn{2}{c}{Wiki + Books $\rightarrow$ PubMed} &  PubMed (half time) & PubMed \\
Vocab & Wiki + Books & PubMed & PubMed & PubMed \\
\midrule
BC5-chem     &   92.85 &  \textbf{93.41} &   93.05 &   93.33 \\
BC5-disease  &   84.70 &  85.43 &   85.02 &   \textbf{85.62} \\
NCBI-disease &   \textbf{89.13} &  87.60 &   87.77 &   87.82 \\
BC2GM        &   83.82 &  84.03 &   84.11 &   \textbf{84.52} \\
JNLPBA       &   78.55 &  79.01 &   78.98 &   \textbf{79.10} \\
\specialrule{0.05pt}{1.5pt}{1.5pt}
EBM PICO     &   73.18 &  \textbf{73.80} &   73.74 &   73.38 \\
\specialrule{0.05pt}{1.5pt}{1.5pt}
ChemProt     &   76.14 &  77.05 &   76.69 &   \textbf{77.24} \\
DDI          &   80.88 &  81.96 &   81.21 &   \textbf{82.36} \\
GAD          &   82.36 &  82.47 &   82.8 &   \textbf{83.96} \\
\specialrule{0.05pt}{1.5pt}{1.5pt}
BIOSSES      &   89.52 &  89.93 &   92.12 &   \textbf{92.30} \\
\specialrule{0.05pt}{1.5pt}{1.5pt}
HoC          &   81.54 &  \textbf{83.14} &   82.13 &   82.32 \\
\specialrule{0.05pt}{1.5pt}{1.5pt}
PubMedQA     &   \textbf{60.24} &  54.84 &   55.28 &   55.84 \\
BioASQ       &   84.14 &  79.00 &   79.43 &   \textbf{87.56} \\
\specialrule{0.05pt}{1.5pt}{1.5pt}
BLURB score  &   80.34 &  80.03 &   80.23 &   \textbf{81.16} \\
\bottomrule
\end{tabular}
\end{center}
\caption{Evaluation of the impact of pretraining corpora and time on the performance on BLURB. In the first two columns, pretraining was first conducted on Wiki \& Books, then on PubMed abstracts. All use the same amount of compute (twice as long as original BERT pretraining), except for the third column, which only uses half (same as original BERT pretraining).}
\label{tab:ablation-corpora}
\end{table}

\begin{table}[ht]
\begin{center}
\begin{tabular}{lccc}
\specialrule{1pt}{1.5pt}{1.5pt}
&  PubMed  &  PubMed + PMC & PubMed + PMC (longer training) \\
\midrule
BC5-chem         &   93.33 &   \textbf{93.36} &      93.34 \\
BC5-disease      &   85.62 &   85.62 &      \textbf{85.76} \\
NCBI-disease     &   87.82 &   \textbf{88.34} &      88.04 \\
BC2GM            &   \textbf{84.52} &   84.39 &      84.37 \\
JNLPBA           &   79.10 &   78.90 &      \textbf{79.16} \\
\specialrule{0.05pt}{1.5pt}{1.5pt}
EBM PICO         &   73.38 &   73.64 &      \textbf{73.72} \\
\specialrule{0.05pt}{1.5pt}{1.5pt}
ChemProt         &   \textbf{77.24} &   76.96 &      76.80 \\
DDI              &   82.36 &   \textbf{83.56} &      82.06 \\
GAD              &   83.96 &   \textbf{84.08} &      82.90 \\
\specialrule{0.05pt}{1.5pt}{1.5pt}
BIOSSES          &   92.30 &   90.39 &      \textbf{92.31} \\
\specialrule{0.05pt}{1.5pt}{1.5pt}
HoC              &   82.32 &   82.16 &      \textbf{82.62} \\
\specialrule{0.05pt}{1.5pt}{1.5pt}
PubMedQA         &   55.84 &   \textbf{61.02} &      60.02 \\
BioASQ           &   \textbf{87.56} &   83.43 &      87.20 \\
\specialrule{0.05pt}{1.5pt}{1.5pt}
BLURB score      &   81.16 &   81.01 &      \textbf{81.50} \\
\specialrule{1pt}{1.5pt}{1.5pt}
\end{tabular}
\end{center}
\caption{Evaluation of the impact of pretraining text on the performance of PubMedBERT on BLURB. The first result column corresponds to the standard PubMedBERT pretrained using PubMed abstracts (``PubMed''). The second one corresponds to PubMedBERT trained using both PubMed abstracts and PMC full text (``PubMed+PMC''). The last one corresponds to PubMedBERT trained using both PubMed abstracts and PMC full text, for 60\% longer (``PubMed+PMC (longer training)''). }
\label{tab:pubmedbert-text}
\end{table}


To assess the impact of pretraining options on downstream applications, we conduct several ablation studies using PubMedBERT as a running example. 
\autoref{tab:ablation-vocab-wwm} shows results assessing the effect of vocabulary and whole-word masking (WWM). 
Using the original BERT vocabulary derived from Wikipedia \& BookCorpus (by continual pretraining from the original BERT), the results are significantly worse than using an in-domain vocabulary from PubMed. Additionally, WWM leads to consistent improvement across the board, regardless of the vocabulary in use. 
A significant advantage in using an in-domain vocabulary is that the input will be shorter in downstream tasks, as shown in \autoref{tab:input-length}, which makes learning easier. \autoref{fig:attention-examples} shows examples of how domain-specific pretraining with in-domain vocabulary helps correct errors from mixed-domain pretraining.  

Furthermore, we found that pretraining on general-domain text provides no benefit even if we use the in-domain vocabulary; see \autoref{tab:ablation-corpora}.
The first column corresponds to BioBERT, which conducted pretraining first on the general domain and then on PubMed. The second column adopted the same continual pretraining strategy, except that the in-domain vocabulary (from PubMed) was used, which actually led to slight degradation in performance. On the other hand, by conducting pretraining from scratch on PubMed, we attained similar performance even with half of the compute (third column), and attained significant gain with the same amount of compute (fourth column; PubMedBERT). In sum, general-domain pretraining confers no advantage here in domain-specific pretraining.  

In our standard PubMedBERT pretraining, we used PubMed abstracts only. We also tried adding full-text articles from PubMed Central (PMC),\footnote{\url{https://www.ncbi.nlm.nih.gov/pmc/}} with the total pretraining text increased substantially to 16.8 billion words (107 GB). Surprisingly, this generally leads to a slight degradation in performance across the board. However, by extending pretraining for 60\% longer (100K steps in total), the overall results improve and slightly outperform the standard PubMedBERT using only abstracts. The improvement is somewhat mixed across the tasks, with some gaining and others losing. We hypothesize that the reason for this behavior is two-fold. First, PMC inclusion is influenced by funding policy and differs from general PubMed distribution, and full texts generally contain more noise than abstracts. As most existing biomedical NLP tasks are based on abstracts, full texts may be slightly out-domain compared to abstracts. Moreover, even if full texts are potentially helpful, their inclusion requires additional pretraining cycles to make use of the extra information.

\begin{table}[ht]
\begin{center}
\begin{tabular}{lcc}
\specialrule{1pt}{1.5pt}{1.5pt}
     &  PubMedBERT &  + adversarial\\
\midrule
BC5-chem     &    \textbf{93.33} & 93.17 \\
BC5-disease  &    \textbf{85.62} & 85.48 \\
NCBI-disease &    87.82 & \textbf{87.99} \\
BC2GM        &    \textbf{84.52} & 84.07 \\
JNLPBA       &    79.10 & \textbf{79.18} \\
\specialrule{0.05pt}{1.5pt}{1.5pt}
EBM PICO     &    \textbf{73.38} & 72.92 \\
\specialrule{0.05pt}{1.5pt}{1.5pt}
ChemProt     &    \textbf{77.24} & 77.04 \\
DDI          &    82.36 & \textbf{83.62} \\
GAD          &    \textbf{83.96} & 83.54 \\
\specialrule{0.05pt}{1.5pt}{1.5pt}
BIOSSES      &    92.30 & \textbf{94.11} \\
\specialrule{0.05pt}{1.5pt}{1.5pt}
HoC          &    \textbf{82.32} & 82.20 \\
\specialrule{0.05pt}{1.5pt}{1.5pt}
PubMedQA     &    \textbf{55.84} & 53.30 \\
BioASQ       &    \textbf{87.56} & 82.71 \\
\specialrule{0.05pt}{1.5pt}{1.5pt}
BLURB score  &    \textbf{81.16} & 80.77 \\
\specialrule{1pt}{1.5pt}{1.5pt}
\end{tabular}
\end{center}
\caption{Comparison of PubMedBERT performance on BLURB using standard and adversarial pretraining. \label{tab:adversarial}}
\end{table}


Adversarial pretraining has been shown to be highly effective in boosting performance in general-domain applications \cite{liu2020alum}. We thus conducted adversarial pretraining in PubMedBERT and compared its performance with standard pretraining (\autoref{tab:adversarial}). Surprisingly, adversarial pretraining generally leads to a slight degradation in performance, with some exceptions such as sentence similarity (BIOSSES). We hypothesize that the reason may be similar to what we observe in pretraining with full texts. Namely, adversarial training is most useful if the pretraining corpus is more diverse and relatively out-domain compared to the application tasks. We leave a more thorough evaluation of adversarial pretraining to future work.


\subsection{Ablation Study on Fine-Tuning Methods}

\begin{table}[ht]
\begin{center}
\begin{tabular}{lcc}
\specialrule{1pt}{1.5pt}{1.5pt}
Task-Specific Model &  Linear Layer &  Bi-LSTM \\
\midrule
BC5-chem    &         \textbf{93.33} &               93.12 \\
BC5-disease &         85.62 &               \textbf{85.64} \\
JNLPBA      &         \textbf{79.10} &               \textbf{79.10} \\
\specialrule{0.05pt}{1.5pt}{1.5pt}
ChemProt    &         \textbf{77.24} &               75.40 \\
DDI         &         \textbf{82.36} &               81.70 \\
GAD         &         \textbf{83.96} &               83.42 \\
\specialrule{1pt}{1.5pt}{1.5pt}
\end{tabular}
\end{center}
\caption{Comparison of linear layers vs recurrent neural networks for task-specific fine-tuning in named entity recognition (entity-level F1) and relation extraction (micro F1), all using the standard PubMedBERT.}
\label{tab:linear-lstm}
\end{table}

\begin{table}[ht]
\begin{center}
\begin{tabular}{lccc}
\specialrule{1pt}{1.5pt}{1.5pt}
Tagging Scheme &    BIO &  BIOUL &     IO \\
\midrule
BC5-chem    &  93.33 &  \textbf{93.37} &  93.11 \\
BC5-disease &  85.62 &  85.59 &  \textbf{85.63} \\
JNLPBA      &  \textbf{79.10} &  79.02 &  79.05 \\
\specialrule{1pt}{1.5pt}{1.5pt}
\end{tabular}
\end{center}
\caption{Comparison of entity-level F1 for biomedical named entity recognition (NER) using different tagging schemes and the standard PubMedBERT.}
\label{tab:ner-tagging}
\end{table}

In the above studies on pretraining methods, we fix the fine-tuning methods to the standard methods described in \autoref{sec:fine-tuning}. 
Next, we will study the effect of modeling choices in task-specific fine-tuning, by fixing the underlying pretrained language model to our standard PubMedBERT (WWM, PubMed vocabulary, pretrained using PubMed abstracts). 

Prior to the current success of pretraining neural language models, standard NLP approaches were often dominated by sequential labeling methods, such as conditional random fields (CRF) and more recently recurrent neural networks such as LSTM. Such methods were particularly popular for named entity recognition (NER) and relation extraction. 

With the advent of BERT models and the self-attention mechanism, the utility of explicit sequential modeling becomes questionable. The top layer in the BERT model already captures many non-linear dependencies across the entire text span. Therefore, it's conceivable that even a linear layer on top can perform competitively. We find that this is indeed the case for NER and relation extraction, as shown in \autoref{tab:linear-lstm}. The use of a bidirectional LSTM (Bi-LSTM) does not lead to any substantial gain compared to linear layer.

We also investigate the tagging scheme used in NER. The standard tagging scheme distinguishes words by their positions within an entity. 
For sequential tagging methods such as CRF and LSTM, distinguishing the position within an entity is potentially advantageous compared to the minimal IO scheme that only distinguishes between inside and outside of entities. But for BERT models, once again, the utility of more complex tagging schemes is diminished. We thus conducted a head-to-head comparison of the tagging schemes using three biomedical NER tasks in BLURB. 
As we can see in \autoref{tab:ner-tagging}, the difference is minuscule, suggesting that with self-attention, the sequential nature of the tags is less essential in NER modeling.

\begin{table}[ht]
\begin{center}
\begin{tabular}{llll}
\specialrule{1pt}{1.5pt}{1.5pt}
Input text &  Classification Encoding    &   ChemProt    &   DDI \\
\midrule
ENTITY DUMMIFICATION    &[CLS]  &         77.24 &               82.36 \\
ENTITY DUMMIFICATION    &MENTION    &    77.22 &               82.08 \\
ORIGINAL        &[CLS]  &         50.52 &               37.00 \\
ORIGINAL        &MENTION    &        75.48 &               79.42 \\
ENTITY MARKERS  &[CLS]  &     \textbf{77.72} &               82.22 \\
ENTITY MARKERS  &MENTION  &        77.22 &                \textbf{82.42} \\
ENTITY MARKERS  &ENTITY START   &         77.58 &                82.18 \\
\bottomrule
\end{tabular}
\end{center}
\caption{Evaluation of the impact of entity dummification and relation encoding in relation extraction, all using PubMedBERT. With entity dummification, the entity mentions in question are anonymized using entity type tags such as \$DRUG or \$GENE. With entity marker, special tags marking the start and end of an entity are appended to the entity mentions in question. Relation encoding is derived from the special $\tt [CLS]$ token appended to the beginning of the text or the special entity start token, or by concatenating the contextual representation of the entity mentions in question. 
}
\label{tab:rel-repr}
\end{table}

The use of neural methods also has subtle, but significant, implications for relation extraction. Previously, relation extraction was generally framed as a classification problem with manually-crafted feature templates. To prevent overfitting and enhance generalization, the feature templates would typically avoid using the entities in question. Neural methods do not need hand-crafted features, but rather use the neural encoding of the given text span, including the entities themselves. This introduces a potential risk that the neural network may simply memorize the entity combination. This problem is particularly pronounced in self-supervision settings, such as distant supervision, because the positive instances are derived from entity tuples with known relations. As a result, it is a common practice to ``dummify'' entities (i.e., replace an entity with a generic tag such as \$DRUG or \$GENE)~\cite{wang18emnlp,jia19naacl}. 

This risk remains in the standard supervised setting, such as in the tasks that comprise BLURB. 
We thus conducted a systematic evaluation of entity dummification and relation encoding, using two relation extraction tasks in BLURB.

For entity marking, we consider three variants: dummify the entities in question; use the original text; add start and end tags to entities in question.
For relation encoding, we consider three schemes. In the $\tt [CLS]$ encoding introduced by the original BERT paper, the special token $\tt [CLS]$ is prepended to the beginning of the text span, and its contextual representation at the top layer is used as the input in the final classification. Another standard approach concatenates the BERT encoding of the given entity mentions, each obtained by applying max pooling to the corresponding token representations. 
Finally, following prior work, we also consider simply concatenating the top contextual representation of the entity start tag, if the entity markers are in use \cite{soares2019matching}. 

\autoref{tab:rel-repr} shows the results. Simply using the original text indeed exposes the neural methods to significant overfitting risk. Using $\tt [CLS]$ with the original text is the worst choice, as the relation encoding has a hard time to distinguish which entities in the text span are in question. Dummification remains the most reliable method, which works for either relation encoding method. Interestingly, 
using entity markers leads to slightly better results in both datasets, as it appears to prevent overfitting while preserving useful entity information. We leave it to future work to study whether this would generalize to all relation extraction tasks.

\section{Discussion}
\label{sec:discussion}






Standard supervised learning requires labeled examples, which are expensive and time-consuming to annotate. Self-supervision using unlabeled text is thus a long-standing direction for alleviating the annotation bottleneck using transfer learning. Early methods focused on clustering related words using distributed similarity, such as Brown Clusters~\cite{brown1992class,liang2005semi}. 
With the revival of neural approaches, neural embedding has become the new staple for transfer learning from unlabeled text. This starts with simple stand-alone word embeddings~\cite{mikolov2013efficient,pennington2014glove}, and evolves into more sophisticated pretrained language models, from LSTM in ULMFiT~\cite{howard2018ulm-ft} and ELMo~\cite{peters2018elmo} to transformer-based models in GPT~\cite{radford2018improving,radford2019language} and BERT~\cite{devlin2018bert,liu2019roberta}. Their success is fueled by access to large text corpora, advanced hardware such as GPUs, and a culmination of advances in optimization methods, such as Adam~\cite{kingma2014adam} and slanted triangular learning rate~\cite{howard2018ulm-ft}.
Here, transfer learning goes from the pretrained language models to fine-tuning task-specific models for downstream applications. 

As the community ventures beyond the standard newswire and Web domains, and begins to explore high-value verticals such as biomedicine, a different kind of transfer learning is brought into play by combining text from various domains in pretraining language models. The prevailing assumption is that such mixed-domain pretraining is advantageous. In this paper, we show that this type of transfer learning may not be applicable when there is a sufficient amount of in-domain text, as is the case in biomedicine. In fact, our experiments comparing clinical BERTs with PubMedBERT on biomedical NLP tasks show that even related text such as clinical notes may not be helpful, since we already have abundant biomedical text from PubMed. Our results show that we should distinguish different types of transfer learning and separately assess their utility in various situations.

There are a plethora of biomedical NLP datasets, especially from various shared tasks such as BioCreative~\cite{smith2008overview,Arighi2011,Mao2014OverviewOT,Kim2015OverviewOB}, BioNLP~\cite{10.5555/2107691.2107693,DBLP:conf/bionlp/2019}, SemEval~\cite{DBLP:conf/semeval/2018,DBLP:conf/semeval/2017,DBLP:conf/semeval/2016,DBLP:conf/semeval/2013}, and BioASQ~\cite{nentidis2019results}. 
The focus has evolved from simple tasks, such as named entity recognition, to more sophisticated tasks, such as relation extraction and question answering, and new tasks have been proposed for emerging application scenarios such as evidence-based medical information extraction~\cite{nye2018corpus}. 
However, while comprehensive benchmarks and leaderboards are available for the general domains (e.g., GLUE~\cite{wang19iclr_glue} and SuperGLUE~\cite{wang2019superglue}), they are still a rarity in biomedical NLP. In this paper, inspired by prior effort towards this direction~\cite{peng2019transfer}, we create the first leaderboard for biomedical NLP, BLURB --- a comprehensive benchmark containing thirteen datasets for six tasks. 

\section{Conclusion}
\label{sec:conclusion}

In this paper, we challenge a prevailing assumption in pretraining neural language models and show that domain-specific pretraining from scratch can significantly outperform mixed-domain pretraining such as continual pretraining from a general-domain language model, leading to new state-of-the-art results for a wide range of biomedical NLP applications. 
To facilitate this study, we create BLURB, a comprehensive benchmark for biomedical NLP featuring a diverse set of tasks such as named entity recognition, relation extraction, document classification, and question answering. To accelerate research in biomedical NLP, we release our state-of-the-art biomedical BERT models and setup a leaderboard based on BLURB.

Future directions include: further exploration of domain-specific pretraining strategies; incorporating more tasks in biomedical NLP; extension of the BLURB benchmark to clinical and other high-value domains.

\bibliographystyle{ACM-Reference-Format}
\bibliography{sample-base,ref_xiaodong,ref_datasets,ref_mlucas,ref_hoifung,ref_tristan,ref_rob}










\end{document}